\documentclass[conference]{IEEEtran}
\IEEEoverridecommandlockouts
\usepackage{cite}
\usepackage{amsmath,amssymb,amsfonts}
\usepackage{algorithmic}
\usepackage{graphicx}
\usepackage{textcomp}
\usepackage{xcolor}
\usepackage{url}
\usepackage{subfig}
\usepackage{multirow}
\usepackage{multicol}
\def\BibTeX{{\rm B\kern-.05em{\sc i\kern-.025em b}\kern-.08em
    T\kern-.1667em\lower.7ex\hbox{E}\kern-.125emX}}
\begin{document}

\title{SkySeg: Collaborative Onboard Semantic Segmentation with Heterogeneous UAVs in the Wild\\
% {\footnotesize \textsuperscript{*}Note: Sub-titles are not captured in Xplore and
% should not be used}
% \thanks{Identify applicable funding agency here. If none, delete this.}
}

\author{\IEEEauthorblockN{1\textsuperscript{st} Anqi Lu}
\IEEEauthorblockA{\textit{Harbin Institute of Technology} \\
Harbin, China \\
luanqi@stu.hit.edu.cn}
\and
\IEEEauthorblockN{2\textsuperscript{nd} Yun Cheng}
\IEEEauthorblockA{\textit{ETH Zurich} \\
Zurich, Switzerland \\
yun.cheng@sdsc.ethz.ch}
\and
\IEEEauthorblockN{3\textsuperscript{rd} Youbing Hu}
\IEEEauthorblockA{\textit{Huawei Cloud Algorithm Innovation Lab} \\
Be, China \\
huyoubing3@huawei.com}
\and
\IEEEauthorblockN{4\textsuperscript{th} Zhiqiang Cao}
\IEEEauthorblockA{\textit{Harbin Institute of Technology} \\
Harbin, China \\
zhiqiang\_cao@stu.hit.edu.cn}
\and
\IEEEauthorblockN{5\textsuperscript{th} Jie Liu}
\IEEEauthorblockA{\textit{Harbin Institute of Technology} \\
Harbin, China \\
mjieliu@outlook.com}
\and
\IEEEauthorblockN{6\textsuperscript{th} Zhijun Li}
\IEEEauthorblockA{\textit{Harbin Institute of Technology} \\
Harbin, China \\
lizhijun\_os@hit.edu.cn}
}

\maketitle

\begin{abstract}
The demand for unmanned aerial vehicle (UAV)-based image acquisition and analysis has surged, with UAVs increasingly utilized for semantic segmentation tasks. To meet the real-time analysis requirements of UAV remote sensing missions, performing onboard computation and making decisions based on the results is a natural approach. However, deploying semantic segmentation on resource-constrained UAV platforms presents two significant challenges: 1) hardware constraints limit the ability of UAVs to perform real-time semantic segmentation, and 2) environmental variations during flight cause data distribution shifts, deviating from the original training data. To address these issues, this paper introduces SkySeg, a heterogeneous multi-UAV air-air cooperation framework that integrates computer vision and flight pattern to enable onboard semantic segmentation using low-cost sensors. SkySeg employs an efficient information fusion inference method, combining low-definition, wide-area images with high-definition, focused-area images. Additionally, it incorporates a cross-device test-time adaptation (TTA) strategy to enhance segmentation performance in dynamic environments by collaboratively addressing distribution shifts of test data streams across UAVs. Experimental results demonstrate that our SkySeg framework accelerates inference latency by approximately 3.6x, improves onboard segmentation accuracy by 5.91\%, and achieves a 10.91\% average accuracy gain in the wild. 
% Code is anonymously available at \url{https://anonymous.4open.science/r/SkySeg-0040}.
\end{abstract}


\section{Introduction}

\begin{figure}[t]
\centerline{\includegraphics[width=0.44\textwidth, height=0.48\textwidth]{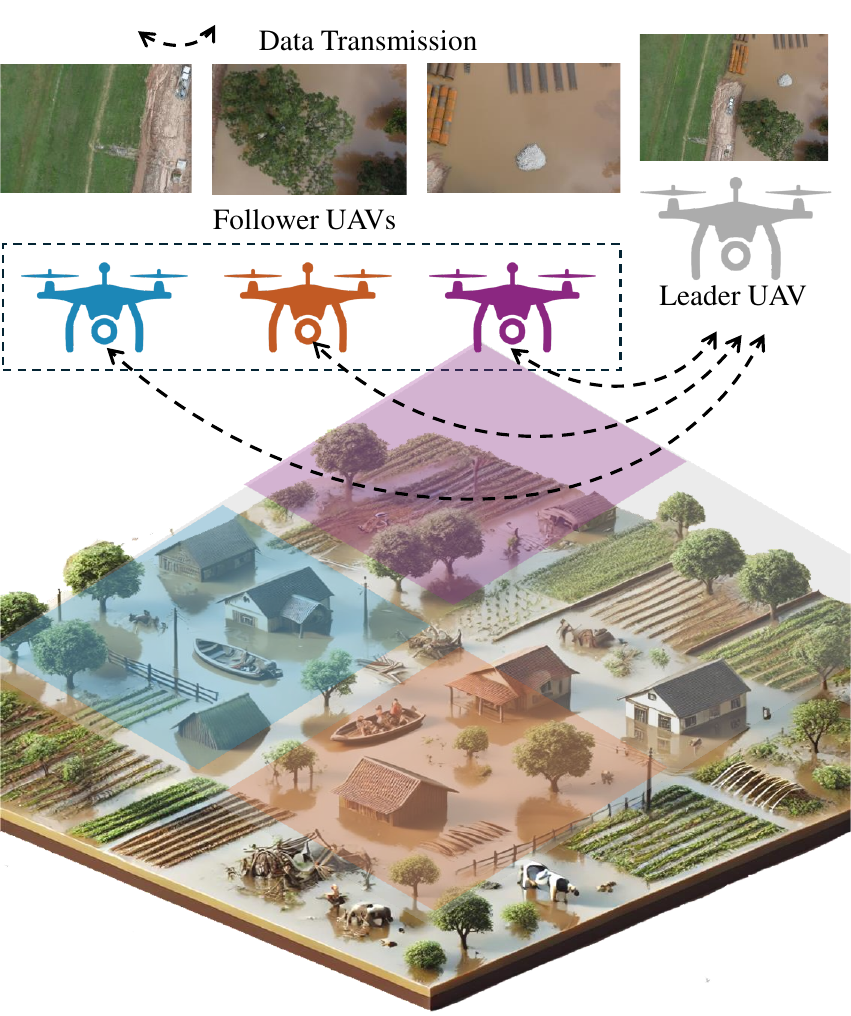}}
\caption{SkySeg uses multi-UAV air-air cooperation to achieve onboard semantic segmentation in disaster monitoring. The leader UAV captures wide-area images at high altitudes, while the follower UAVs capture focused-area images at low altitudes based on guidance from the leader UAV.}
\label{fig:case}
\end{figure}

With the rapid advancement of unmanned aerial vehicle (UAV) technology \cite{xu2025mmuavsense}, UAVs have become indispensable tools in areas like agricultural monitoring \cite{su2022lodgenet} and damage assessment \cite{liang2023uav}. Unlike satellites, UAVs flying at low altitudes offer flexible deployment, rapid response, and cost efficiency, making them ideal for hazardous or labor-intensive tasks such as flood and wildfire monitoring \cite{goudarzi2021real}. These scenarios require precise environmental analysis, highlighting semantic segmentation vital for extracting detailed information from images taken at various angles and altitudes \cite{kemker2018algorithms}. For real-time applications, UAVs must analyze images and provide immediate insights during flight. As autonomous UAV technology evolves, onboard segmentation plays a key role in enabling environment perception and decision-making \cite{nex2022uav}.

However, deploying semantic segmentation on UAVs is challenging due to hardware limitations and dynamic flight environments. UAVs are constrained by \underline{S}ize, \underline{W}eight, \underline{a}nd \underline{{P}}ower (SWaP), limiting memory, computational capacity, and battery life \cite{tran2019energy}. Meanwhile, UAV-captured remote sensing images are often high-resolution and data-intensive. Deep learning models for semantic segmentation, which often rely on deep architectures with many channels for accuracy, are typically too resource-intensive for onboard use \cite{liu2021light}. Moreover, UAVs operate in complex and variable wild, leading to distribution shifts in the test data stream compared to the training data, which negatively impacts model generalization \cite{schneider2020improving}. For instance, weather conditions such as rain, snow, fog, or dust can introduce occlusions and blur images, while sensor degradation may add noise \cite{zhang2023perception}. Therefore, developing adaptive, efficient methods for onboard semantic segmentation remains a critical and challenging research area in UAV applications. 

To tackle these challenges, this paper introduces SkySeg, an edge-assisted heterogeneous multi-UAV cooperation framework that enables efficient onboard semantic segmentation through air-air collaboration. As illustrated in Fig.~\ref{fig:case}, SkySeg demonstrates how multiple UAVs operating at different altitudes collaborate to perform inference in disaster scenarios. Specifically, a leader UAV captures low-definition, wide-coverage panoramic images to provide an initial overview of the entire area. Based on this preliminary information, several follower UAVs focus on acquiring high-definition, localized images of critical regions. By coordinating UAVs flying over affected areas, SkySeg ensures rapid and accurate environmental analysis. Once critical regions are identified by the leader UAV, the follower UAVs conduct detailed scans and perform semantic segmentation onboard, significantly improving response speed and decision-making effectiveness.

The SkySeg framework combines computer vision and flight pattern to facilitate collaborative onboard semantic segmentation. By equipping multiple UAVs with low-cost sensors, SkySeg enables sensing and inference tasks through heterogeneous cooperation, effectively overcoming the resource constraints of individual UAVs. Leveraging their high mobility, UAVs can quickly reach target areas for data collection guided by Global Positioning System (GPS). SkySeg employs a fusion inference strategy that integrates low-definition, wide-area images with high-definition, focused-area images, striking a balance between inference speed and segmentation accuracy to enhance overall performance. To adapt to wild environments, test-time adaptation (TTA) is a lightweight solution by adapting model parameters during testing without requiring source data \cite{wang2020tent}. SkySeg incorporates a cross-device TTA mechanism, enabling the multi-UAV system to adapt to environmental changes in real time based on the current test data stream during flight. It effectively mitigates the problem of data distribution shifts caused by environmental variations, enhancing the robustness and reliability of semantic segmentation.

Overall, the main contributions of this paper are as follows.

\begin{itemize}
\item We present a heterogeneous multi-UAV collaborative framework that integrates computer vision and flight pattern, utilizing low-cost sensors to achieve real-time and accurate onboard semantic segmentation in the wild. To the best of our knowledge, this is the first work to combine collaborative perception with cross-device TTA.
\item We design an onboard information fusion inference method for semantic segmentation, where the leader UAV integrates low-definition, wide-area image results with high-definition, focused-area image results from the follower UAVs. This approach enables efficient semantic segmentation of aerial remote sensing data.
\item We propose a cross-device TTA method under multi-UAV cooperation to improve the adaptability of onboard semantic segmentation in UAV flight wild.
\item Experimental results on public UAV remote sensing datasets demonstrate that the SkySeg framework enhances segmentation accuracy and reduces inference latency while maintaining robust performance in the wild.
\end{itemize}

\section{Related Work}
\subsection{Semantic Segmentation on UAVs}
Traditional UAV remote sensing systems primarily focus on image collection, with processing performed on ground-based computing devices \cite{olson2021review}. However, real-time tasks are hindered by limited network bandwidth, prompting a shift toward onboard computation for faster decision-making and efficient result transmission \cite{deng2020lightweight}. This requires semantic segmentation models that balance efficiency and accuracy under hardware constraints. Research on lightweight UAV semantic segmentation is emerging, with most relying on convolutional neural networks (CNNs) \cite{cheng2024methods}. 
MAVNet \cite{nguyen2019mavnet} is developed a real-time semantic segmentation model for micro aerial vehicles, providing two task-specific datasets, to meet SWaP constraints.
A lightweight weed mapping network architecture \cite{deng2020lightweight} is designed to integrate map visualization, flight control, image capture, and real-time processing.
An efficient and lightweight UAV semantic segmentation model \cite{liu2021light} is proposed to incorporate attention mechanisms to capture global pixel relationships and feature correlations. 
RTSDM \cite{li2022rtsdm} is a real-time semantic mapping system supporting power-limited UAV navigation by integrating semantic segmentation for key SLAM frames.
LSMA \cite{shen2023semantic} is a lightweight sea scene semantic segmentation method using DeepLabV3+ \cite{chen2018encoder} with parallel multi-scale attention feature fusion.
A lightweight CNN-Transformer hybrid architecture \cite{lu2024lightweight} is developed to feature a decoder with query-value squeeze axial attention for handling multi-scale objects efficiently.

\subsection{Multi-UAV Cooperation}
A single UAV faces limitations in coverage, field of view, endurance, and computational resources, making it challenging to handle complex tasks. To address this, research increasingly focuses on multi-UAV collaboration.
SkyStitch \cite{meng2015skystitch} enhances video surveillance by stitching video streams from multiple UAVs, using distributed feature extraction, flight controller hints, and Kalman filter-based state estimation to improve stitching speed and quality.
A search experimental platform \cite{wang2021design} utilizes a three-UAV leader-follower team for image-based coordinated target search, addressing autonomous cooperative control challenges.
A search and rescue system \cite{xing2022multi} leverages YOLOv5 and multi-UAV collaboration to reduce energy consumption and costs in wilderness search and rescue, enhancing computational efficiency and object detection.
SkyNet \cite{peng2023skynet} implements multi-UAV searching to overcome poor visibility and resource constraints, enabling real-time human identification and localization with task scheduling between edge devices and cloud servers to reduce delays.
Air-CAD \cite{tan2024air} combines air-ground collaboration to address single UAV limitations, improving crowd anomaly detection accuracy and reducing inference delays through multi-feature analysis.

The above multi-UAV cooperation methods address single UAV limitations effectively. However, no research has yet explored multi-UAV collaboration for efficient onboard semantic segmentation infield, balancing low latency and high accuracy.

\begin{figure*}[th]
	\centering
        \subfloat[Segmentation accuracy.]{\includegraphics[width=0.32\textwidth]{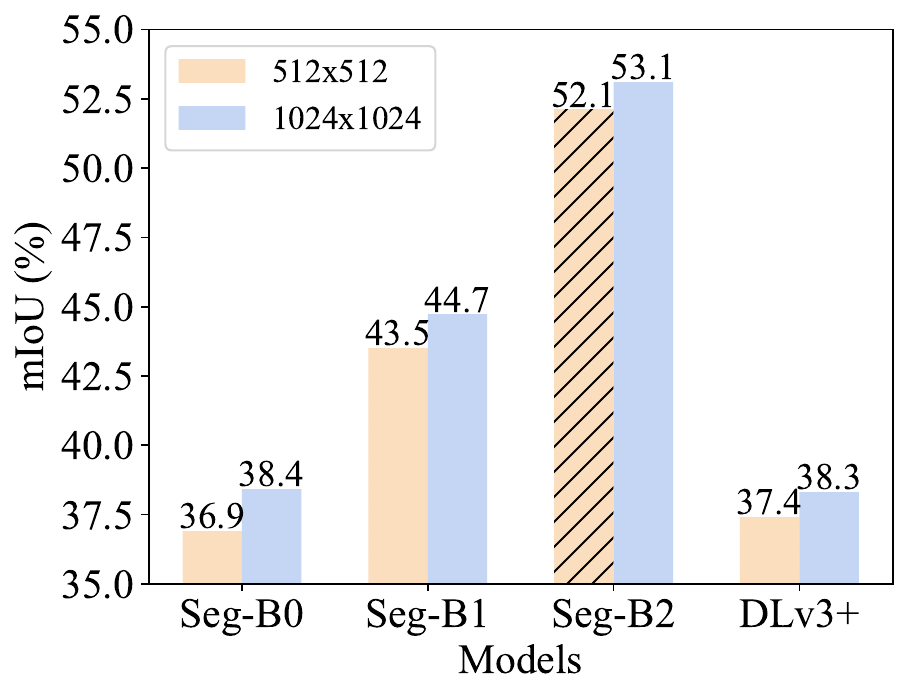}%
        }  
        \hfil
        \subfloat[Computational overhead.]{\includegraphics[width=0.32\textwidth]{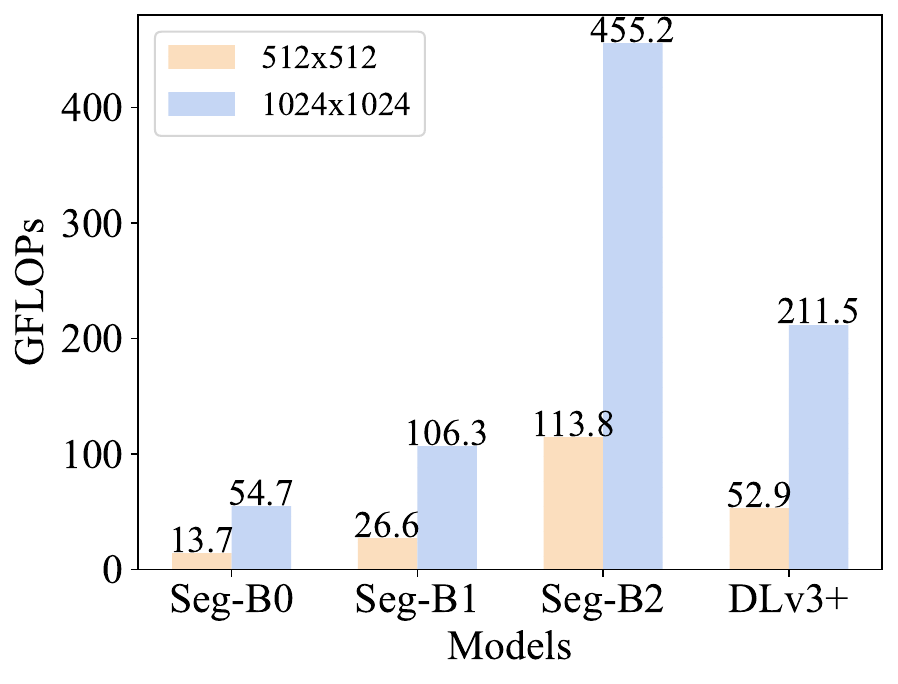}%
        }
        \hfil
        \subfloat[Inference latency.]{\includegraphics[width=0.32\textwidth]{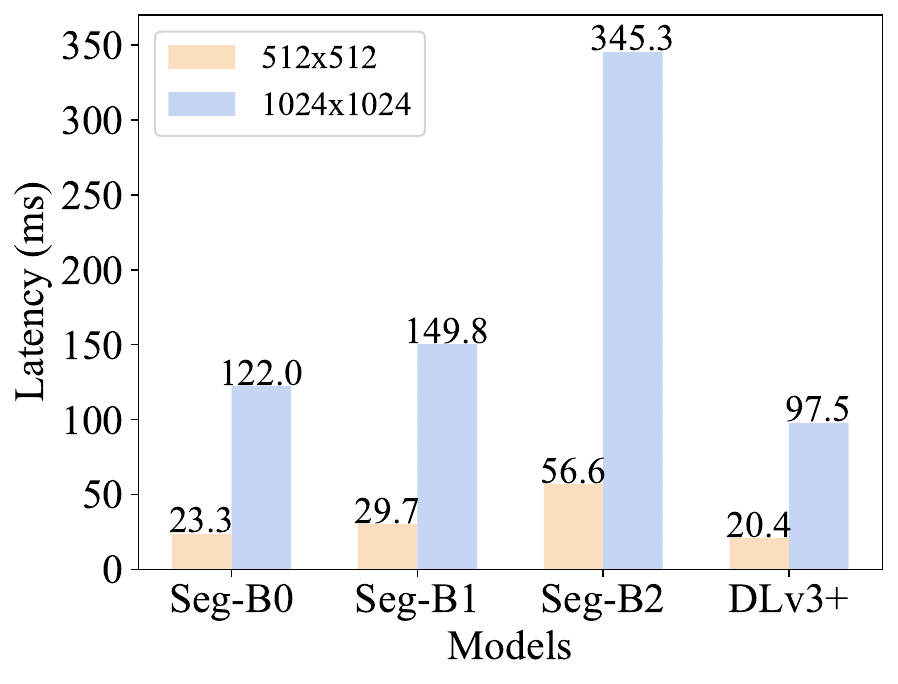}%
        }
        \quad
        \caption{Impact of different models with different input resolutions on the SDD. As the input resolution increases, all three metrics increase accordingly, with computational overhead and inference latency rising dramatically.}
	\label{fig:motivation1}
\end{figure*}

\section{Motivational Studies}
We assess how a single UAV affects segmentation accuracy, computational latency, and mean Intersection-over-Union (mIoU) \cite{guo2018review} performance. Experiments are run on a desktop with an Intel i9-9900K CPU and an NVIDIA RTX 2080Ti GPU. We test two segmentation models: the Transformer-based SegFormer \cite{xie2021segformer} and the CNN-based DeepLabv3+ \cite{chen2018encoder} with the Semantic Drone Dataset (SDD) \cite{sdd2020}.

\subsection{Impact on Performance of Semantic Segmentation} 
\textbf{Size.} Due to the limited storage of edge devices, we compare SegFormer-B0, B1, B2, and DeepLabv3+ with MobileNet-V2 as the backbone, with model sizes of 7.5 MB, 27.4 MB, 54.8 MB, and 11.8 MB, respectively.

\textbf{mIoU.} Fig.~\ref{fig:motivation1}(a) depicts the segmentation accuracy of different semantic segmentation models with 512 $\times$ 512 and 1024 $\times$ 1024 input resolutions. Higher resolution improves accuracy slightly, with gains under 2\%.

\textbf{FLOPs.} Fig.~\ref{fig:motivation1}(b) shows the computational overhead of various segmentation models with different input resolutions. The result shows that the input resolution of 1024 $\times$ 1024 is 4 times more computationally intensive than 512 $\times$ 512.

\textbf{Latency.} Fig.~\ref{fig:motivation1}(c) illustrates the inference latency of different segmentation models at different input resolutions. Latency increases sharply with resolution, exceeding 5 times at 1024 $\times$ 1024 compared to 512 $\times$ 512.

\textbf{Insight 1.} \textit{In dense prediction tasks such as semantic segmentation, image resolution has a more pronounced effect on computational overhead and inference latency than model size. While downsampling input resolution can substantially enhance inference speed, it comes at the expense of segmentation accuracy.}

\subsection{Impact on Performance in the Wild}
\begin{figure}[t]
	\centering
        \includegraphics[width=0.35\textwidth]{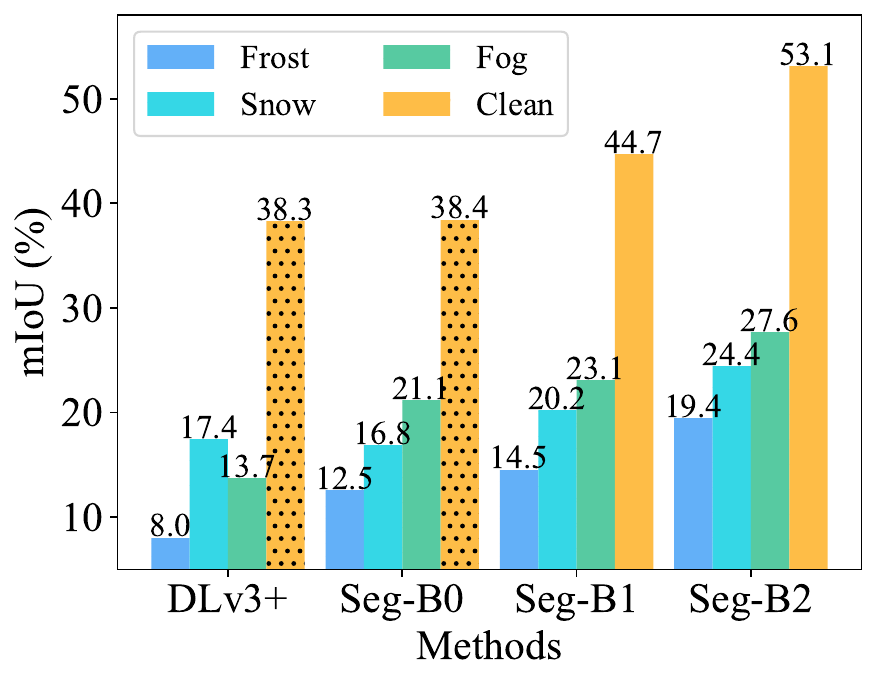}
	\caption{Impact of different models on segmentation accuracy with the input resolution of 1024 $\times$ 1024 on the SDD. Segmentation accuracy decreases dramatically in dynamic environments (e.g., snow, fog, frost).}
	\label{fig:motivation2}
\end{figure}

\textbf{mIoU.} Fig.~\ref{fig:motivation2} presents the segmentation performance of various trained models in snow, fog, and frost (severity level 5) at 1024 $\times$ 1024 input resolution. Weather-induced degradation causes a significant performance drop, averaging over 20\% compared to clean test data.

\textbf{Insight 2.} \textit{Environmental factors in the wild, such as weather conditions, lighting variations, and sensor degradation, can impact data collection and image quality, leading to a divergence between the test data distribution and the original training data. Consequently, trained models often experience significant performance degradation when applied to semantic segmentation on newly collected images.}

\subsection{Summary and Motivation}
The results above highlight the significant impact of image resolution and the data collection environment on semantic segmentation. Regarding image resolution, high-resolution images require more computational resources to process each pixel. To enhance inference speed, downsampling to a lower resolution is commonly used. However, this approach can result in a loss of accuracy, as the model may miss fine-grained details, affecting segmentation quality. Regarding the data collection environment, real-world UAV flights are subject to unpredictable environmental changes. Weather variations, such as rain, snow, fog, or sand, can introduce occlusions into the images. Trained models typically learn features from data within a specific distribution, but when test data distribution shifts significantly, the model's generalization ability is compromised. Without exposure to these new "shifted" conditions during training, the model lacks adaptability.

Considering these factors, we design SkySeg as follows: 1) Employing multiple UAVs to capture critical details in low-resolution images, and 2) Leveraging statistics from newly collected data across multiple UAVs to enhance the model's adaptability in the wild.

\section{Overview of SkySeg}
We present SkySeg, a heterogeneous multi-UAV collaboration framework designed for efficient onboard semantic segmentation in the wild.

\textbf{Low-cost sensors.} High-resolution image processing demands significant computational resources. Existing methods often downsample or crop images, sacrificing detail or context and reducing segmentation accuracy \cite{zhang2023review}. Higher resolutions also require more GPU memory \cite{li2024memory}. To address this, SkySeg uses low-cost sensors on each UAV to capture low-resolution images, cutting costs and computation. By combining information from multiple UAVs, SkySeg reconstructs high-resolution segmentation outputs.

\textbf{Multi-UAV formation control.} SkySeg adopts a leader-follower formation \cite{villa2020outdoor}. The leader UAV has stronger computing capabilities and guides follower UAVs by providing GPS coordinates. Follower UAVs maintain relative positions and collect data accordingly \cite{wang2021design}. Since the focus is on segmentation, UAV flight control details are not discussed.

\begin{figure}[t]
	\centering	
       \includegraphics[width=0.43\textwidth]{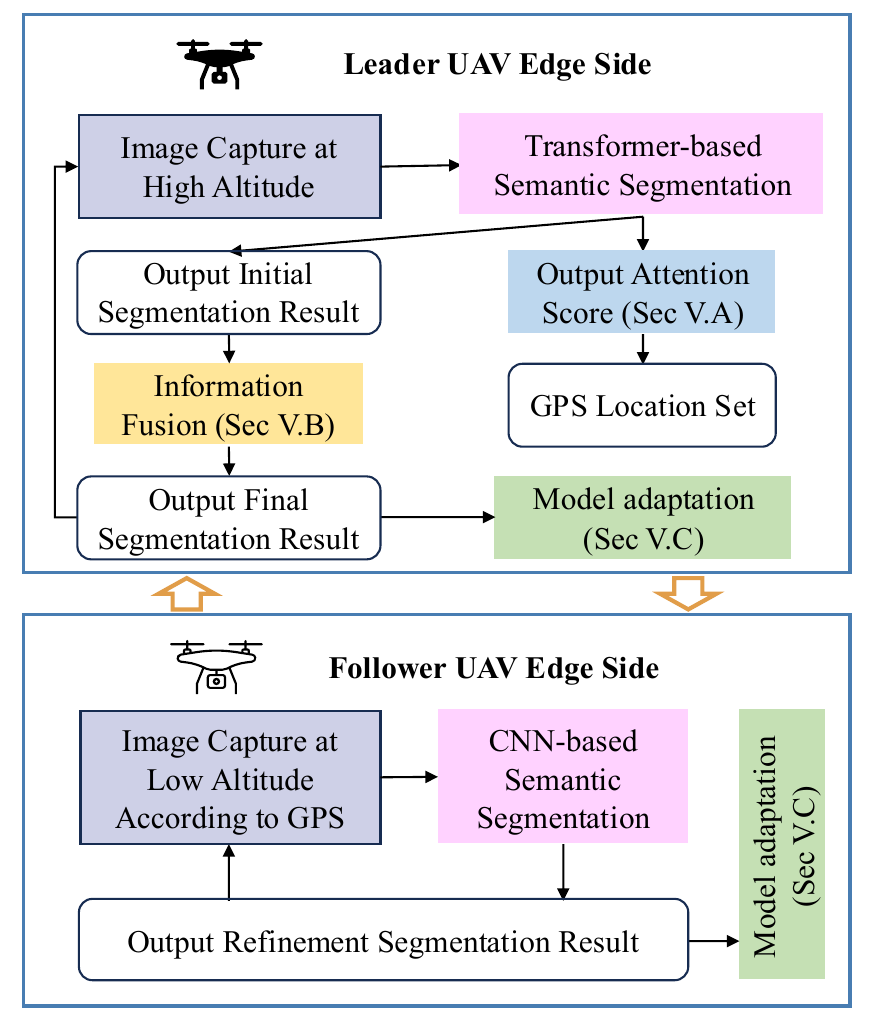}
	\caption{The operational flow of SkySeg, which starts with the leader UAV and then all the follower UAVs work in parallel.}
	\label{fig:framework}
\end{figure}

\textbf{Workflow.} As shown in Fig.~\ref{fig:framework}, the leader UAV flies at high altitude, collecting wide-area low-definition images and performing initial segmentation using a Transformer-based model. It identifies key regions needing refinement and sends their coordinates to the followers. Follower UAVs fly lower, capture high-definition images of those regions, and use a CNN-based model for fine-grained segmentation. Their results are sent back to the leader for fusion into the final output. To adapt in the wild, the leader updates its model using new image statistics, while followers share and aggregate statistics to update theirs. This process repeats, continuously improving segmentation accuracy.

\begin{figure*}[t]
	\centering	            
        \includegraphics[width=0.75\textwidth]{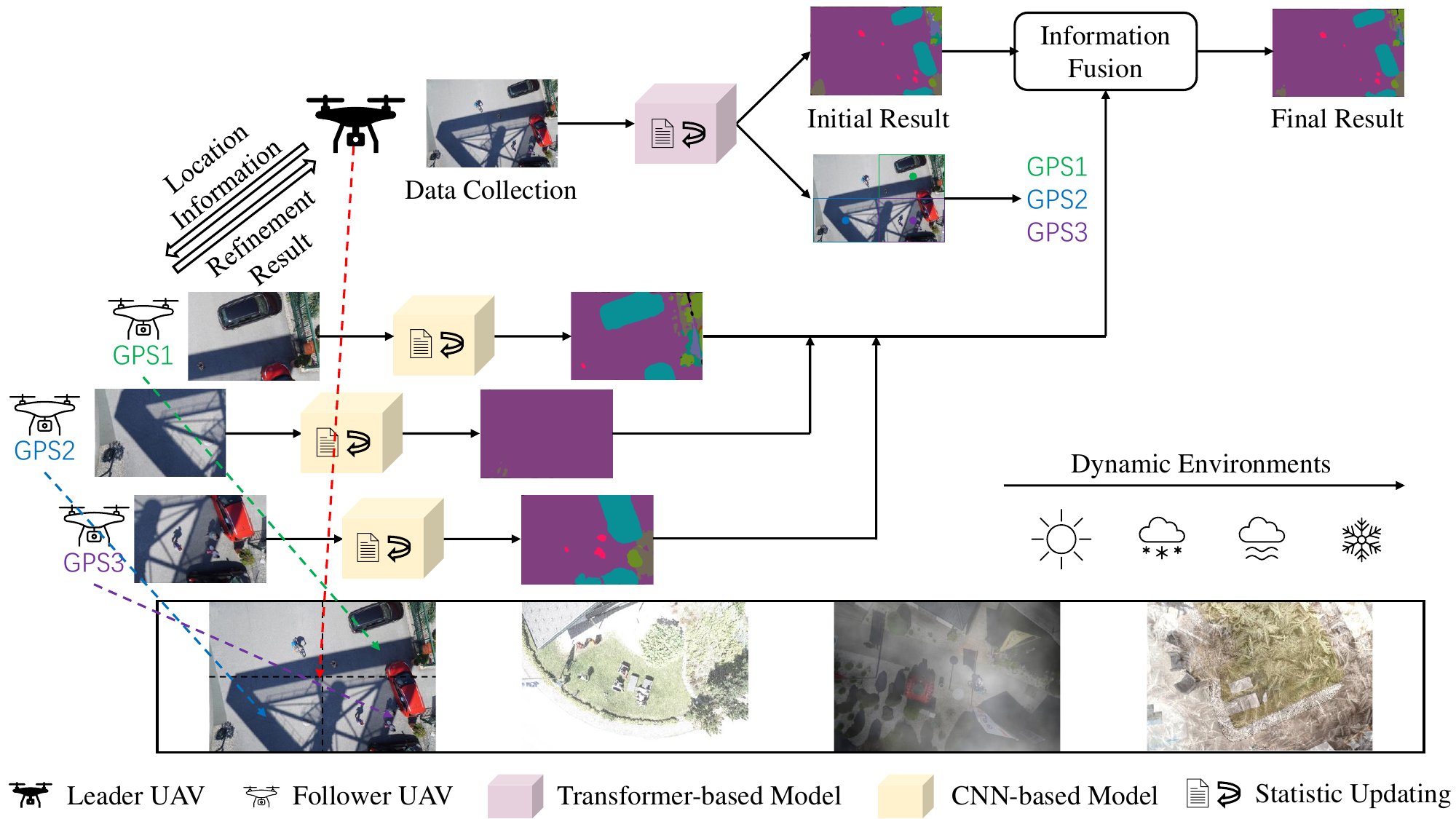}
	\caption{SkySeg is deployed in the case of a leader UAV and three follower UAVs working in a dynamic environment.}
	\label{fig:frameworkcase}
\end{figure*}

\textbf{Case.} Based on the above SkySeg's workflow , we give a detailed example of the SkySeg framework running in a real environment, as shown in Fig.~\ref{fig:frameworkcase}. This case consists of a leader UAV and three follower UAVs. In addition, the real environment is dynamically changing, and the environment may experience from a bright sunny day to a snowy day, etc., when UAVs collect perception data. The leader UAV determines the image patches that need to be improved through onboard real-time analysis of the acquired data. It sends their corresponding GPS information to the three follower UAVs. All the follower UAVs complete the data acquisition and onboard semantic segmentation tasks in parallel and return the results to the leader UAV. The leader UAV finally gets the final result for further decision making.

\textbf{Problem definition.} 
We define the UAV set as $U = \left\{ U_L,U_F^1,\ldots,U_F^n \right\}$, where $U_L$ is the leader UAV and $U_F^1,\ldots,U_F^n$ are follower UAVs. The leader $U_L$ captures low-definition images $I_L \in \mathcal{R}_{low}^{H \times W}$ from high altitude, while the follower $U_F^i$ captures high-definition images $I_F^i \in \mathcal{R}_{high}^{H \times W}$ from low altitude. $H$ and $W$ denote the height and width of the image. During the inference phase, the leader utilizes a Transformer-based model $M_{Transformer}(I_L)$ to generate preliminary results $R_L$. The follower uses a CNN-based model $M_{CNN}(T_F^i)$ to produce detailed results $R_F^i$. To guide refinement, the leader extracts attention scores $A_L = Attention(M_{Transformer}(I_L))$ and extracts relevant area coordinates $C_L = Extraction(A_L)$. The goal is to maximize the final high-resolution fused segmentation result by adapting: the leader’s model using Layer Normalization (LN) statistics $Adapt_{LN}(R_L)$ and the followers’ models using Batch Normalization (BN) statistics $Adapt_{BN}(R_F^{i=1,\ldots,n})$, formulated as:

\begin{equation}
max(Fusion(Adapt_{LN}(R_L), Adapt_{BN}(R_F^{i=1,\ldots,n})))
\end{equation}

\section{Detailed Design of SkySeg}
\subsection{Attention-Based Image Patch Selection}
UAVs with low-cost sensors capture low-resolution images, but upsampling them can lead to detail loss. To address this, SkySeg analyzes wide-area images on the leader UAV to find key regions that need more detail.

\begin{figure}[t]
	\centering	
        \includegraphics[width=0.48\textwidth]{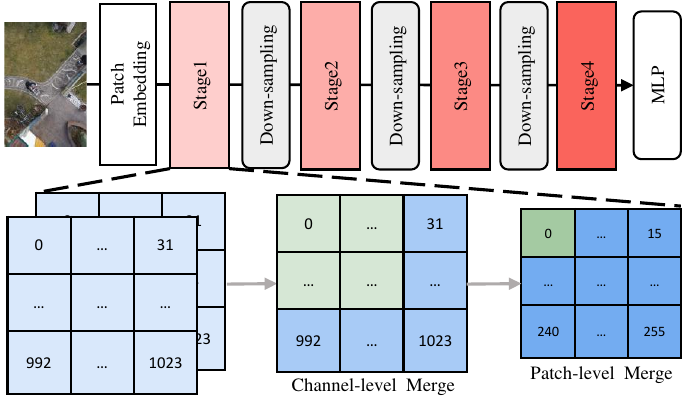}
	\caption{Attention-based image patch selection method. 
    % In stage 1, attention scores are extracted by first averaging image patches at the channel level, then creating single-channel feature maps at the patch level. The process is similar for other stages. Finally, attention scores from all four stages are weighted and summed.
    }
	\label{fig:attention}
\end{figure}

A Transformer-based model is used to generate attention scores for image patches, helping identify important areas for refinement. To suit the limited computing power of UAVs, we design a lightweight method, as illustrated in Fig.~\ref{fig:attention}.

The model adopts a hierarchical feature extraction approach, capturing multi-scale information through progressive downsampling and convolution operations. As shown in Fig.~\ref{fig:attention}, the model includes four stages. For a 512 $\times$ 512 $\times$ 3 input, the PatchEmbed layer divides the image into 128 $\times$ 128 patches and embeds them into 64 dimensions. Each stage has two Transformer blocks, producing attention maps of 128 $\times$ 128, 64 $\times$ 64, 32 $\times$ 32, and 16 $\times$ 16. Attention is computed as:

\begin{equation}
A = Softmax(\frac{Q K^\text{T}}{\sqrt{d}})
\end{equation}
where $Q$ and $K$ are the \textit{query} and \textit{key} matrices, and $d$ is the dimension \cite{chen2023diffrate} of \textit{query/key}. The fusion process of multi-scale attention is as follows.

\textit{Channel-level merge.} Average attention maps from the two Transformer blocks in each stage can be given by 

\begin{equation}
A_k = \frac{1}{2}(A_{k,1} + A_{k,2})
\end{equation}
where $A_k$ denotes the attention map obtained at stage $k$, $k \in \{1,2,3,4\}$. $A_{k,1}$ and $A_{k,2}$ represent the attention maps from the first and second Transformer blocks at stage $k$.

\textit{Patch-level merge.} We apply averaging with window sizes of 8 $\times$ 8, 4 $\times$ 4, and 2 $\times$ 2 to the attention maps from the first three stages to obtain 16 $\times$ 16 maps. The resulting attention maps for the four stages are expressed as $A_1^{\textquotesingle}$, $A_2^{\textquotesingle}$, $A_3^{\textquotesingle}$, and $A_4$.

\textit{Weighted sum.} 
We combine the four 16 $\times$ 16 maps into a fused map $A_{fuse}$ is

\begin{equation}
A_{fuse} = {\omega}_1 \cdot A_1^{\textquotesingle} + {\omega}_2 \cdot A_2^{\textquotesingle} + {\omega}_3 \cdot A_3^{\textquotesingle} + {\omega}_4 \cdot A_4
\end{equation}
where ${\omega}_1$, ${\omega}_2$, ${\omega}_3$, and ${\omega}_4$ are the weighting factors of the four stages. To generate the final 2 $\times$ 2 attention map $A_{final}$, we apply an 8 $\times$ 8 averaging window to the fused attention map.

The leader UAV uses the attention map $A_{final}$ to select top-scoring patches and sends their GPS coordinates to follower UAVs. 
Since the segmentation model is shared, this selection method adds no extra computation or storage cost.

\subsection{Information Fusion}
\label{subsec:detail}
To enable high-resolution segmentation, SkySeg fuses coarse segmentation from low-definition wide-area images with fine segmentation from high-definition focused patches.

\begin{figure}[t]
	\centering	
        \includegraphics[width=0.48\textwidth]{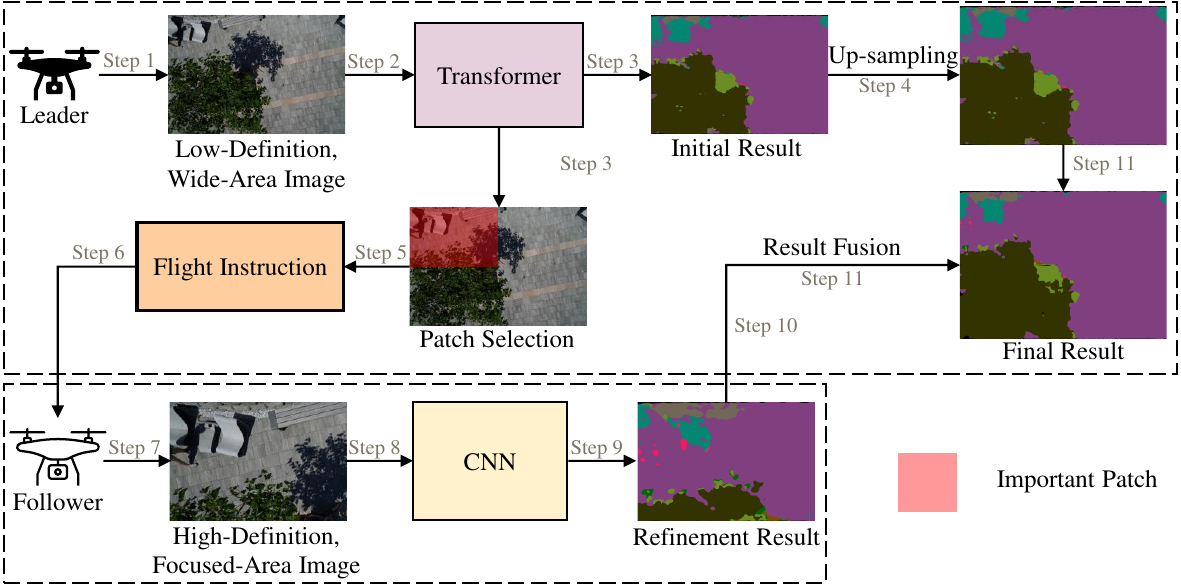}
	\caption{The workflow of the "low-definition + high-definition" information fusion method through the collaboration of high-altitude and low-altitude flight patterns.}
	\label{fig:fusion}
\end{figure}

SparseRefine \cite{liu2025sparse} introduces an innovative "dense + sparse" paradigm for efficient high-resolution segmentation, which enhances dense low-resolution predictions with a sparse set of pixels with the lowest confidence. Inspired by this approach, SkySeg adopts a multi-UAV cooperative framework, where fine-grained predictions of critical image patches enhance coarse-grained predictions. Fig.~\ref{fig:fusion} illustrates the "low-definition + high-definition" paradigm.

\textit{Leader UAV.} It captures a low-definition wide-area image (Step 1) and uses a Transformer-based model for initial segmentation and attention score generation (Step 2, 3). The coarse result is upsampled to a higher resolution (Step 4), and attention scores help identify key areas needing refinement (Step 5). The leader then sends these coordinates to follower UAVs (Step 6). Once it receives refined results, it fuses them with the coarse result using two information fusion methods (Step 11). \textbf{Replacement Fusion:} Refined areas replace coarse ones. \textbf{Probability Comparison Fusion:} Compares class probabilities pixel by pixel and selects the higher one.

\textit{Follower UAV.} It flies to the given GPS coordinates, captures high-definition images (Step 7), performs segmentation using a CNN-based model (Step 8), and generates refinement results (Step 9). It then sends refined pixel-level class predictions and probabilities back to the leader UAV (Step 10).

SkySeg combines the strengths of Transformer (global context) and CNN (local detail) models to achieve accurate segmentation. Onboard semantic segmentation tasks require robust models to handle dynamic and unpredictable conditions. SegFormer and DeepLabv3+ have demonstrated strong performance under common corruptions and perturbations \cite{xie2021segformer,kamann2020benchmarking}.

\subsection{Cross-device TTA}

\begin{figure}[t]
	\centering	            
        \includegraphics[width=0.4\textwidth]{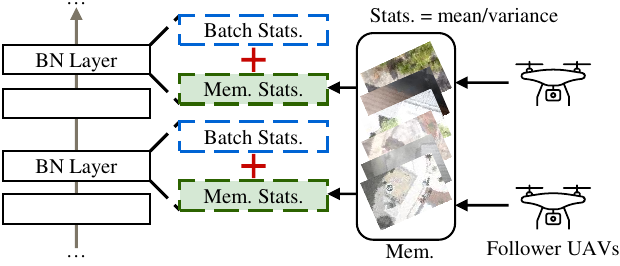}
	\caption{Method for cross-device TTA. The “+” indicates that the model is adapted using the statistic estimation method of the exponential moving average.}
	\label{fig:TTA}
\end{figure}

To address UAV hardware constraints, we use onboard model adaptation by updating statistical parameters without changing model weights. In SkySeg, the leader UAV applies the LN Adaptation method, while follower UAVs implement a cross-device TTA approach (see Fig.~\ref{fig:TTA}).

For the leader UAV's Transformer-based model, we apply LN Adaptation to adjust the LN layer mean and variance at test time. Unlike BN, LN normalizes each sample independently, without batch-specific statistics. Given an input feature map $X$ ($C \times H \times W$, where $C$ is the number of channels, $H$ and $W$ are the height and width of the feature map), the running mean $\mu_L$ and variance $\sigma_L^2$ from the $t$-th batch are computed across the channel and spatial dimensions as

\begin{equation}
\mu_L(t) = \frac{1}{C \times H \times W} \sum_{c=1}^{C} \sum_{a=1}^{H} \sum_{b=1}^{W} X_{c,a,b}
\end{equation}

\begin{equation}
\sigma_L^2(t) = \frac{1}{C \times H \times W} \sum_{c=1}^{C} \sum_{a=1}^{H} \sum_{b=1}^{W} (X_{c,a,b} - \mu_L)^2
\end{equation}

We use the exponential moving average (EMA) \cite{chiley2019online, yuan2023robust, hong2023mecta} method for statistical estimation, the statistic at iteration $t$ update as

\begin{equation}
\hat{\mu_L(t)} = (1-\alpha) \mu_L(t-1) + \alpha \mu_L(t)
\end{equation}

\begin{equation}
\hat{\sigma_L^2(t)} = (1-\alpha) \sigma_L^2(t-1) + \alpha \sigma_L^2(t)
\end{equation}
where the parameter $\alpha \in [0,1]$ controls the memory length. We initialize the statistics using the training data, denoted as $(\mu_L(0),\sigma_L^2(0))$.

For the follower UAV's CNN-based model, we use the BN Adaptation method to dynamically update the BN layer’s mean and variance during testing. In TTA studies for image classification \cite{liang2024comprehensive}, SAR \cite{niu2023towards} highlights that with a batch size of 1, BN statistics may become inaccurate, causing performance degradation in online adaptation. Although larger batch sizes improve adaptation, the high input resolution (e.g., 512 $\times$ 512 or higher) in semantic segmentation makes this memory-intensive \cite{ni2024distribution}. Moreover, real-time UAV applications demand rapid BN updates to adapt to dynamic scenes \cite{lan2021real}. Our cross-device TTA method enhances the effective batch size without extra memory overhead by performing TTA on each follower UAV individually (batch size 1) and then averaging BN statistics across UAVs via low-latency communication. Each follower UAV integrates statistics from its training data, local data, and shared data from other UAVs. The statistical calculation is performed separately for each feature map channel. For the input feature map $X$ ($1 \times C \times H \times W$), the running mean $\mu_c^i$ and variance $(\sigma_c^i)^2$ of channel $c$ from the $t$-th batch are computed as

\begin{equation}
\mu_c^i(t) = \frac{1}{H \times W} \sum_{a=1}^{H} \sum_{b=1}^{W} X_{c,a,b}
\end{equation}

\begin{equation}
(\sigma_c^i)^2(t) = \frac{1}{H \times W} \sum_{a=1}^{H} \sum_{b=1}^{W} (X_{c,a,b} - \mu_c^i)^2
\end{equation}
For the follower UAV $U_F^i$, it receives statistics from other follower UAVs and stores them in the memory bank. The model is adapted using the EMA method. The updated statistics at iteration $t$ are

\begin{equation}
\hat{\mu_c^i(t)} = (1-\alpha) \mu_c^i(t-1) + \alpha \sum_{i=1}^{n}\mu_c^i(t)
\end{equation}

\begin{equation}
\hat{(\sigma_c^i)^2(t)} = (1-\alpha) (\sigma_c^i)^2(t-1) + \alpha \sum_{i=1}^{n}(\sigma_c^i)^2(t)
\end{equation}
Similarly, we initialize the statistics for the follower UAV $U_F^i$ using its training data, with the initial values $(\mu_c^i(0),(\sigma_c^i)^2(0))$.

\section{Evaluation}

\subsection{Experimental Setup}
\textbf{Hardware.} We use the NVIDIA Jetson TX2 as the onboard computing device on UAVs. The TX2 has a dual-core NVIDIA Denver2 CPU, a 256-core NVIDIA Pascal GPU, and 8GB of memory, meeting the SWaP requirements for UAVs \cite{cao2020research}.

\textbf{Implementation.} All code is implemented using the PyTorch 3.8 framework \cite{paszke2017automatic}. All models are pre-trained on a high-performance server with a 12-core AMD R9 5900X CPU, 128 GB memory, and a GeForce RTX 3090 GPU. Besides, we set $\alpha$=0.05 \cite{yuan2023robust} in the EMA method.

\textbf{Dataset.} We use two publicly available UAV segmentation datasets to evaluate SkySeg: Semantic Drone Dataset (SDD) \cite{sdd2020} and FloodNet dataset \cite{rahnemoonfar2021floodnet}. SDD focuses on urban scenes with 400 images (6000 $\times$ 4000 px) and 23 classes, resized to 600 $\times$ 400 px for low-resolution processing. FloodNet covers flood-related scenes with 2343 images (4000 $\times$ 3000 px) and 10 classes, resized to 400 $\times$ 300 px. To simulate dynamic conditions, we create corrupted versions—SDD-C and FloodNet-C—by adding weather noise (snow, fog, frost). To assess the impact of different severity levels, SDD-C uses severity level 5, and FloodNet-C uses level 3.

\textbf{Metrics.} We evaluate SkySeg using: 1) mIoU (\%) for segmentation accuracy, 2) Latency (ms) for inference speed, and 3) Data transmission volume for the amount of data sent.

\subsection{Overall Performance}
We evaluate the SkySeg with a multi-UAV setup consisting of one leader UAV and three follower UAVs. 
The performance evaluation of the SkySeg is shown in Table~\ref{tab:overall_SDD} and Table~\ref{tab:overall_FloodNet}.

\begin{table*}[t]\tiny
\caption{Overall Performance Comparison on the SDD-C dataset}
\begin{center}
\resizebox{\linewidth}{!}{
\begin{tabular}{|c|c|c|c|c|c|c|c|}
\hline
Method & Sensor & Latency(ms) & Clean(\%) & Snow(\%) & Fog(\%) & Frost(\%) & Mean(\%) \\
\hline
SegFormer-B1 \cite{xie2021segformer} & High-cost & 1980.011 & 44.71 & 20.17 & 23.05 & 14.51 & 19.24\\
\hline
DeepLabv3+ \cite{chen2018encoder} & High-cost & 837.178 & 38.31 & 17.41 & 13.73 & 7.98 & 13.04\\
\hline
\textbf{SkySeg} & \textbf{Low-cost} & \textbf{546.947*$\downarrow$} & \textbf{50.62$\uparrow$} & \textbf{30.97$\uparrow$} & \textbf{38.14$\uparrow$} & \textbf{21.33$\uparrow$} & \textbf{30.15$\uparrow$}\\
\hline
\end{tabular}
}
\label{tab:overall_SDD}
\end{center}
\end{table*}

\begin{table*}[t]\tiny
\caption{Overall Performance Comparison on the FloodNet-C dataset}
\begin{center}
\resizebox{\linewidth}{!}{
\begin{tabular}{|c|c|c|c|c|c|c|c|}
\hline
Method & Sensor & Latency(ms) & Clean(\%) & Snow(\%) & Fog(\%) & Frost(\%) & Mean(\%) \\
\hline
SegFormer-B1 \cite{xie2021segformer} & High-cost & 1820.024 & 64.20 & 40.27 & 43.21 & 26.06 & 36.51\\
\hline
DeepLabv3+ \cite{chen2018encoder} & High-cost & 760.167 & 65.71 & 36.32 & 37.10 & 25.90 & 33.11\\
\hline
\textbf{SkySeg} & \textbf{Low-cost} & \textbf{497.978*$\downarrow$} & \textbf{69.02$\uparrow$} & \textbf{44.52$\uparrow$} & \textbf{52.61$\uparrow$} & \textbf{31.76$\uparrow$} & \textbf{42.96$\uparrow$}\\
\hline
\end{tabular}
}
\label{tab:overall_FloodNet}
\end{center}
\end{table*}

\textbf{Baseline.} We compare SkySeg with two mainstream semantic segmentation models: SegFormer-B1 \cite{xie2021segformer} and DeepLabv3+ \cite{chen2018encoder}, both deployed on single UAVs with high-cost sensors. 
It is important to note that SkySeg is a general-purpose heterogeneous multi-UAV collaborative framework, rather than a newly designed semantic segmentation model. It offers plug-and-play compatibility, allowing flexible integration of various semantic segmentation methods, all of which can benefit from its collaborative strategies. Due to differences in baseline performance, computational cost, and adaptability, it would be entirely unfair to directly compare it with the latest state-of-the-art (SOTA) models. Therefore, SegFormer-B1 and DeepLabv3+ are chosen as backbone models for SkySeg because (1) they strike a good balance between accuracy and efficiency, making them suitable for UAV deployment; (2) their widespread adoption highlights the generality and practical value of SkySeg; and (3) they demonstrate excellent robustness under common corruptions and perturbations.

\textbf{Inference latency.} Single UAVs experience delays exceeding 800ms on SDD-C and 700ms on FloodNet-C, making them unsuitable for rapid emergency decision-making. In contrast, SkySeg achieves significantly lower latency: 546.947ms on SDD-C (leader UAV: 374.333ms, follower UAVs: 172.614ms) and 497.978ms on FloodNet-C (leader UAV: 344.098ms, follower UAVs: 153.880ms). Compared to the best baselines, SkySeg reduces latency by 290.231ms on SDD-C and 262.189ms on FloodNet-C.

\textbf{Segmentation accuracy.} SkySeg significantly improves segmentation accuracy. On clean datasets, it achieves a 5.91\% improvement on SDD and 3.31\% on FloodNet compared to the best baselines. The gains are even greater on datasets with dynamic environmental corruptions. For SDD-C (severity level 5), accuracy increases by an average of 10.91\%, reaching 15.09\% under foggy conditions. For FloodNet-C (severity level 3), accuracy improves by an average of 6.45\%, reaching 9.4\% under foggy weather. The larger improvement on SDD is due to its lower baseline accuracy and more severe level-5 corruptions, highlighting SkySeg's robustness.

\begin{figure*}[t]
	\centering	            
        \includegraphics[width=\textwidth]{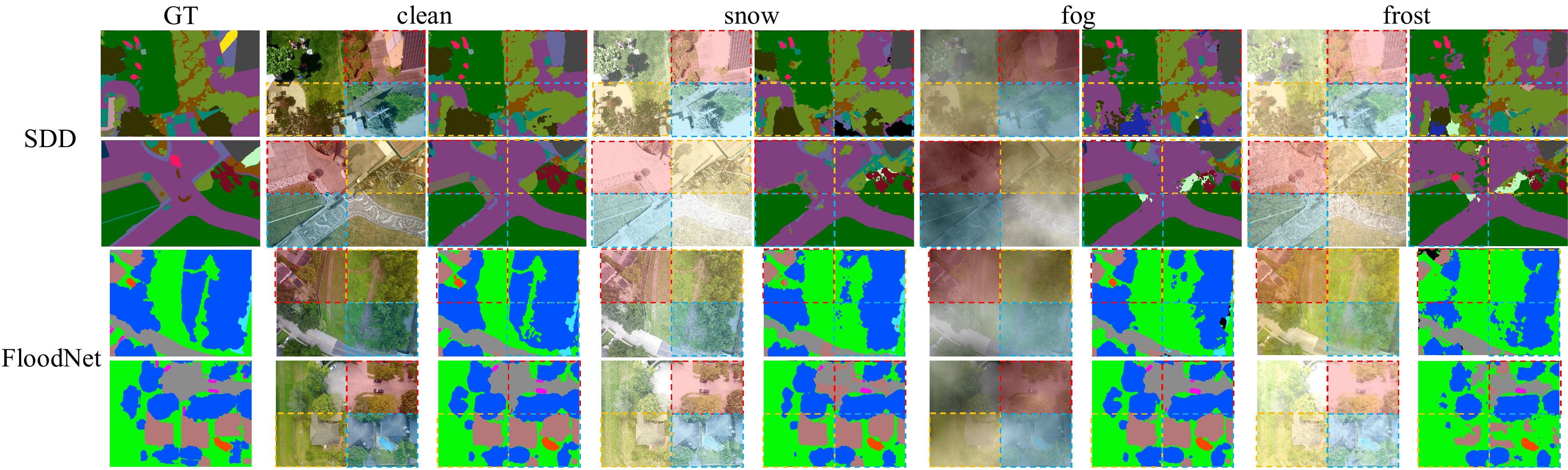}
	\caption{Visualization of SkySeg. SkySeg first identifies the image patches (colored boxes) that need to be improved using an attention-based image patch selection method, then obtains confident predictions of hybrid models based on an information fusion method, and finally ensures the model's adaptability infield using a cross-device TTA method.}
	\label{fig:visual}
\end{figure*}

\textbf{Summary.} SkySeg significantly improves segmentation accuracy across both clean and corrupted datasets, while effectively reducing inference latency and sensor cost. This highlights the framework's efficiency, versatility, and adaptability in the wild. Visualization of SkySeg's results on different datasets are shown in Fig.~\ref{fig:visual}.

\subsection{SkySeg's Module Performance}

\begin{figure}
	\centering
        \subfloat[SDD dataset.]{\includegraphics[width=0.235\textwidth]{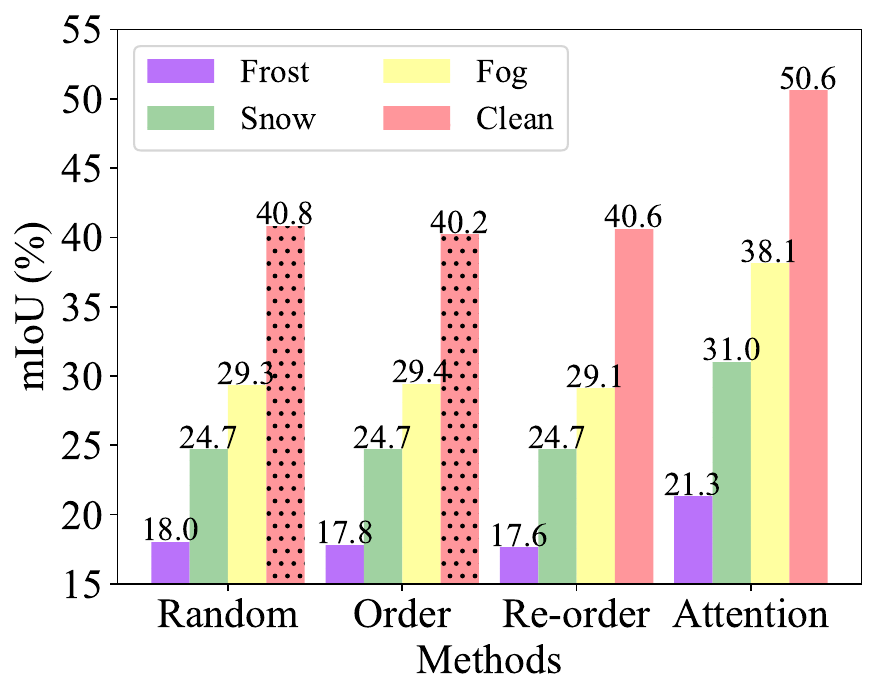}%
        }
        \hfil
        \subfloat[FloodNet dataset.]{\includegraphics[width=0.235\textwidth]{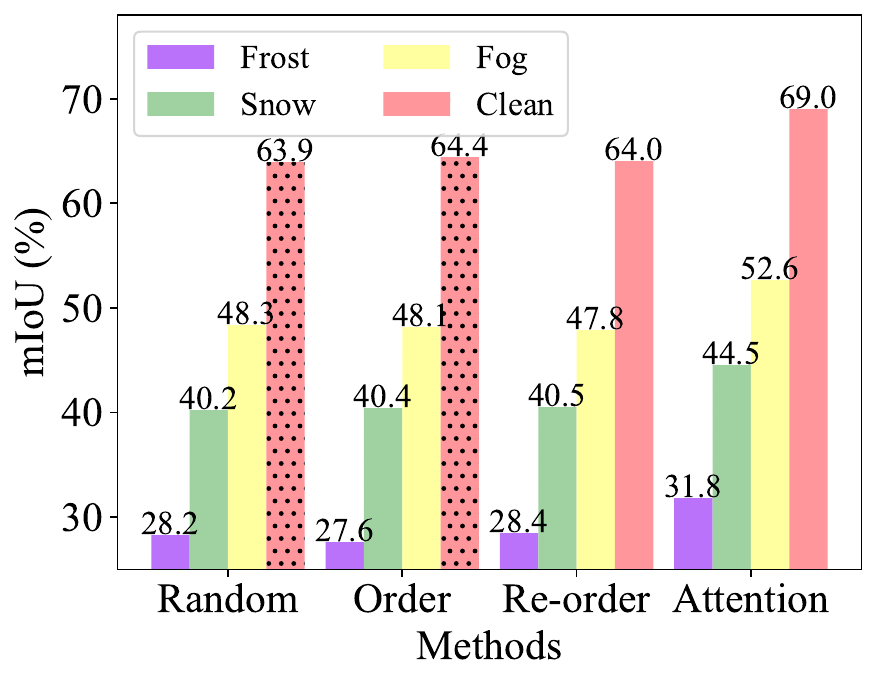}%
        }
        \quad
        \caption{Comparison results of segmentation accuracy under different image patch selection methods.}
	\label{fig:select_result}
\end{figure}

\textbf{Attention-based image patch selection's performance.} We compare four image patch selection methods: random selection (Random), order selection (Order), reverse order selection (Re-order), and our attention-based selection (Attention). Fig.~\ref{fig:select_result} shows the segmentation accuracy across datasets. 
Our attention-based method significantly outperforms others, boosting accuracy on the clean SDD by about 10\% and improving the corrupted dataset average accuracy to 30.15\% (+6\%). On the clean FloodNet, accuracy increases by about 5\%, and on corrupted data, it reaches 42.96\% (+4\%). Besides, this approach shares parameters with the Transformer-based model, adding no extra resource overhead. 
We also assess the effect of weighting factors on segmentation accuracy. The experimental results on different datasets are shown in Table~\ref{tab:weight}. 
Differences are minimal (about 0.1\%), with optimal results achieved when weights are set to ${\omega}_1$=0.1, ${\omega}_2$=0.2, ${\omega}_3$=0.3, and ${\omega}_4$=0.4. It means that prioritizing high-level, fine-grained features in the SegFormer-B1 model enhances patch attention scores, improving segmentation performance.

\begin{table*}\tiny
\caption{Impact of Weighting Factors in Segmentation Accuracy}
\begin{center}
\resizebox{\linewidth}{!}{
\begin{tabular}{|c|c|c|c|c|c|c|c|c|c|}
\hline
%\multirow{2}*{Dataset} & \multirow{2}*{${\omega}_1$} & \multirow{2}*{${\omega}_2$} & \multirow{2}*{${\omega}_3$} & \multirow{2}*{${\omega}_4$} & \multicolumn{5}{c}{mIoU(\%)}\\
%& & & & & Clean & Snow & Fog & Frost & Mean(corrupted)\\
Dataset & ${\omega}_1$ & ${\omega}_2$ & ${\omega}_3$ & ${\omega}_4$ & Clean(\%) & Snow(\%) & Fog(\%) & Frost(\%) & Mean(\%)\\
\hline
\multirow{3}*{SDD-C} & \textbf{0.1} & \textbf{0.2} & \textbf{0.3} & \textbf{0.4} & \textbf{50.62} & \textbf{30.97} & \textbf{38.14} & \textbf{21.33} & \textbf{30.15}\\
\cline{2-10} 
& 0.25 & 0.25 & 0.25 & 0.25 & 50.55 & 30.86 & 38.12 & 20.85 & 29.94 \\
\cline{2-10} 
& 0.4 & 0.3 & 0.2 & 0.1 & 50.27 & 30.09 & 37.67 & 21.20 & 29.65 \\
\hline
\multirow{3}*{FloodNet-C} & \textbf{0.1} & \textbf{0.2} & \textbf{0.3} & \textbf{0.4} & \textbf{69.02} & \textbf{44.52} & \textbf{52.61} & \textbf{31.76} & \textbf{42.96}\\
\cline{2-10}
& 0.25 & 0.25 & 0.25 & 0.25 & 68.84 & 44.16 & 52.51 & 31.58 & 42.75\\
\cline{2-10}
& 0.4 & 0.3 & 0.2 & 0.1 & 68.76 & 44.11 & 52.43 & 31.47 & 42.67\\
\hline
\end{tabular}
}
\label{tab:weight}
\end{center}
\end{table*}

\begin{figure}
	\centering
        \subfloat[SDD dataset.]{\includegraphics[width=0.235\textwidth]{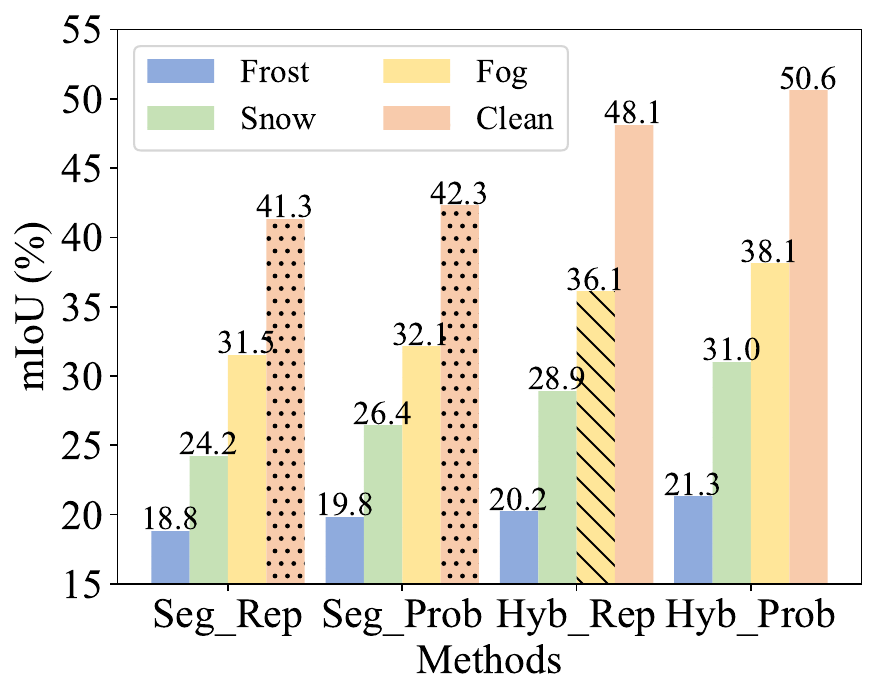}%
        }
        \hfil
        \subfloat[FloodNet dataset.]{\includegraphics[width=0.235\textwidth]{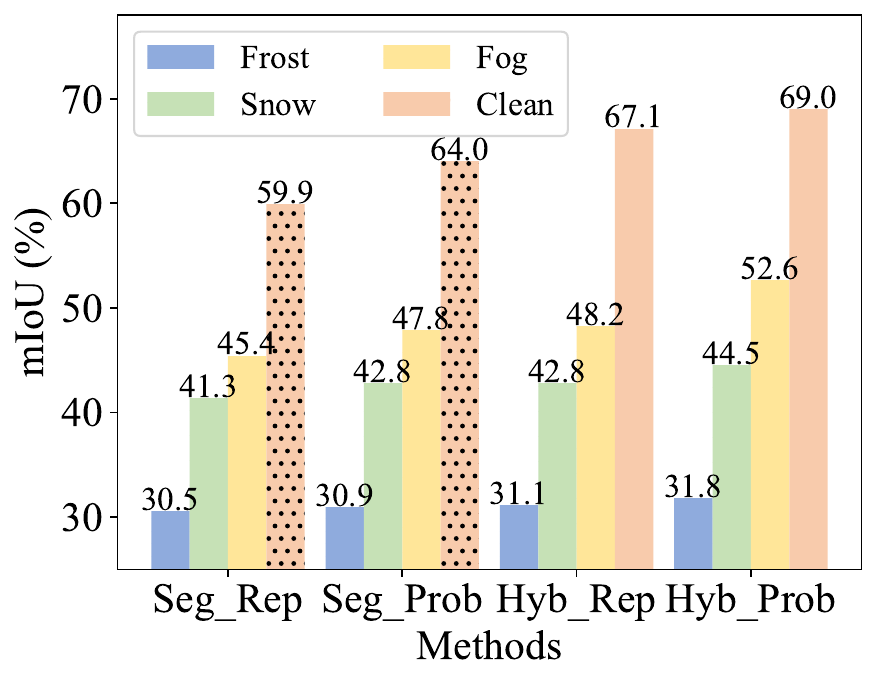}%
        }
        \quad
        \caption{Comparison results of segmentation accuracy under information fusion methods.}
	\label{fig:fusion_result}
\end{figure}

\textbf{Information fusion's performance.} We compare the segmentation accuracy of different information fusion methods: overlay fusion (Seg\_Rep) and maximum probability fusion (Seg\_Prob) using SegFormer-B1 models for all UAVs, and hybrid models—overlay fusion (Hyb\_Rep) and maximum probability fusion (Hyb\_Prob)—which combine Transformer and CNN architectures as described in Section \uppercase\expandafter{\romannumeral5}.B. The results are shown in Fig.~\ref{fig:fusion_result}.
The experiments demonstrate that hybrid models and maximum probability fusion deliver the best segmentation performance. Hybrid models leverage both global features (from Transformers) and local details (from CNNs), while maximum probability fusion further enhances results by integrating the most confident semantic classes.
For SDD, hybrid models improve segmentation accuracy by over 6\% on clean data and over 3\% on corrupted data. For FloodNet, they provide gains of over 5\% on clean data and over 1\% on corrupted data. Additionally, the maximum probability fusion method boosts segmentation accuracy by more than 1\% on both datasets.

\begin{figure}
	\centering
        \subfloat[SDD dataset.]{\includegraphics[width=0.235\textwidth]{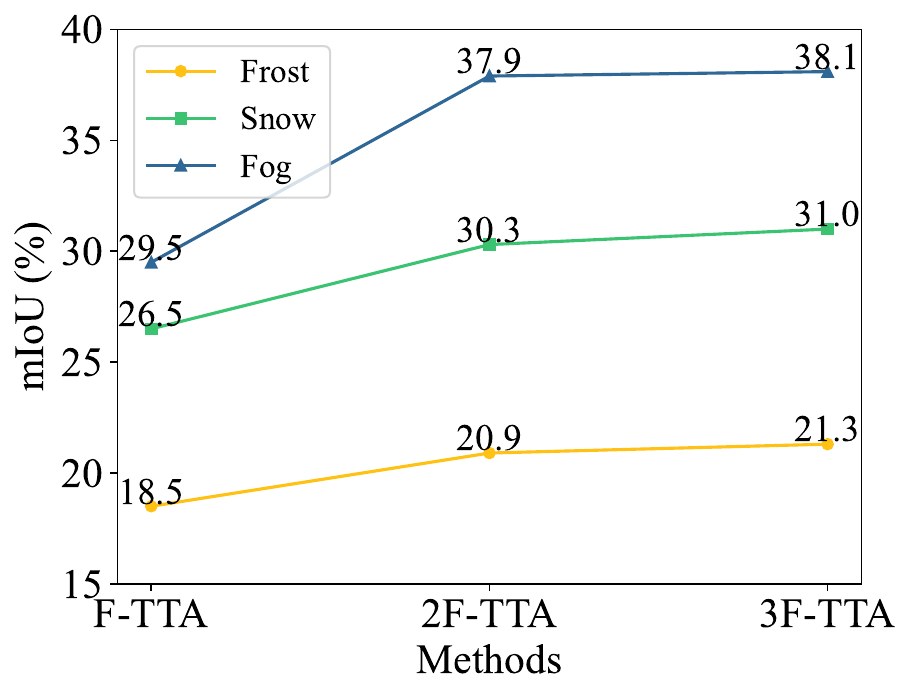}%
        }
        \hfil
        \subfloat[FloodNet dataset.]{\includegraphics[width=0.235\textwidth]{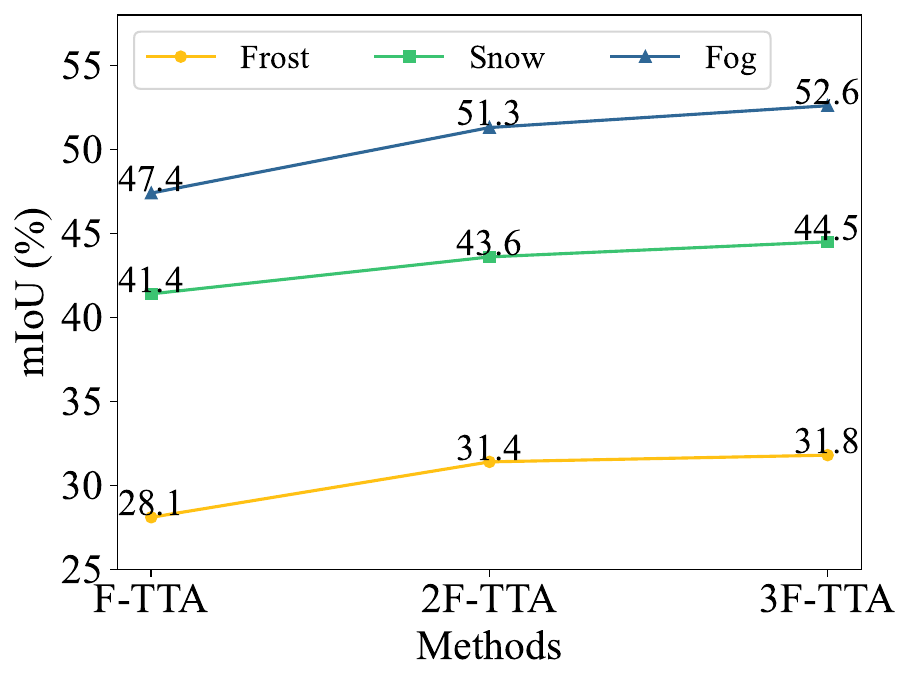}%
        }
        \quad
        \caption{Comparison results of segmentation accuracy under different numbers of UAVs.}
	\label{fig:TTA_acc_result}
\end{figure}

\textbf{Cross-device TTA's performance.} 
We compare the performance of SkySeg with and without cross-device TTA module in Table~\ref{tab:TTA}. Experimental results show that our proposed cross-device TTA can well improve the segmentation performance in the wild.
We compare segmentation accuracy and data transmission volume of cross-device TTA methods with varying numbers of UAVs: one follower UAV (F-TTA), two follower UAVs (2F-TTA), and three follower UAVs (3F-TTA). Segmentation accuracy results are shown in Fig.~\ref{fig:TTA_acc_result}.
Results from both datasets show that segmentation accuracy improves as the number of UAVs increases, with the most significant gain observed when increasing from 1 to 2 devices. 
This improvement stems from BN layers, which normalize data using batch mean and variance. With more follower UAVs, larger sample sizes enhance the model’s stability and adaptability. 
On SDD-C, 2F-TTA improves average accuracy by 4.84\% over F-TTA, while 3F-TTA adds another 0.46\%. On FloodNet-C, 2F-TTA outperforms F-TTA by 3.13\%, with 3F-TTA adding 0.85\%. However, accuracy gains diminish as the number of UAVs continues to increase. 
Regarding data transmission, only the leader UAV uses the Transformer-based model and updates locally without transmission. F-TTA involves one follower UAV, updated entirely locally with no transmission required. For 2F-TTA, each of the two follower UAVs transmits 71.488KB of data to the other. For 3F-TTA, each follower UAV transmits 142.976KB (71.488KB $\times$ 2) to the other two follower UAVs.

\subsection{SkySeg’s Impact on Different Numbers of UAVs}

Three configurations are tested: one leader UAV with one follower (L-F), one leader with two followers (L-2F), and one leader with three followers (L-3F). Table~\ref{tab:framework} summarizes their segmentation accuracy and transmission volume. 

\begin{table}\tiny
\caption{Comparison with and without Cross-Device TTA}
\begin{center}
\resizebox{\linewidth}{!}{
\begin{tabular}{|c|c|c|c|c|}
\hline
Dataset & TTA & Snow(\%) & Fog(\%) & Frost(\%)\\
\hline
\multirow{2}*{SDD-C} & w/o & 21.92 & 25.25 & 17.51\\
\cline{2-5}
& w/ & 30.97 & 38.14 & 21.33\\
\hline
\multirow{2}*{FloodNet-C} & w/o & 41.41 & 45.45 & 27.88\\
\cline{2-5}
& w/ & 44.52 & 52.61 & 31.76\\
\hline
\end{tabular}
}
\label{tab:TTA}
\end{center}
\end{table}

\begin{table*}\tiny
\caption{Framework Performance Evaluation}
\begin{center}
\resizebox{\linewidth}{!}{
\begin{tabular}{|c|c|c|c|c|c|c|c|c|}
\hline
%\multirow{2}*{Dataset} & \multirow{2}*{Method} & \multicolumn{5}{c}{mIoU(\%)} & \multirow{2}*{Fusion Data(MB)} & \multirow{2}*{TTA Data(KB)} \\
%& & Clean & Snow & Fog & Frost & Mean(corrupted) & &\\
Dataset & Method & Clean(\%) & Snow(\%) & Fog(\%) & Frost(\%) & Mean(\%) & Fusion Data(MB) & TTA Data(KB)\\
\hline
\multirow{3}*{SDD-C} & L-F & 45.56 & 26.72 & 29.82 & 19.37 & 25.30 & 1.4 & 0\\
\cline{2-9}
& L-2F & 47.00 & 28.92 & 34.50 & 20.93 & 28.12 & 1.4 & 71.488\\
\cline{2-9}
& L-3F & 50.62 & 30.97 & 38.14 & 21.33 & 30.15 & 1.4 & 142.976\\
\hline
\multirow{3}*{FloodNet-C} & L-F & 65.90 & 44.06 & 49.16 & 30.71 & 41.31  & 0.72 & 0\\
\cline{2-9}
& L-2F & 67.55 & 44.30 & 51.57 & 31.08 & 42.32 & 0.72 & 71.488\\
\cline{2-9}
& L-3F & 69.02 & 44.52 & 52.61 & 31.76 & 42.96 & 0.72 & 142.976\\
\hline
\end{tabular}
}
\label{tab:framework}
\end{center}
\end{table*}

Table~\ref{tab:framework} shows that the segmentation accuracy improves with more follower UAVs, on both clean and corrupted datasets. The data transmission volumes listed in the table are calculated per follower UAV. For information fusion, each follower UAV transmits only refinement segmentation results to the leader UAV, keeping data transmission volume constant regardless of the number of follower UAVs. However, for cross-device TTA, as the number of UAVs increases, follower UAVs must share BN layer statistics with more devices, causing a steady rise in transmission volume.

\begin{figure*}[ht]
	\centering
        \subfloat[SegFormer-B1.]{\includegraphics[width=0.24\textwidth]{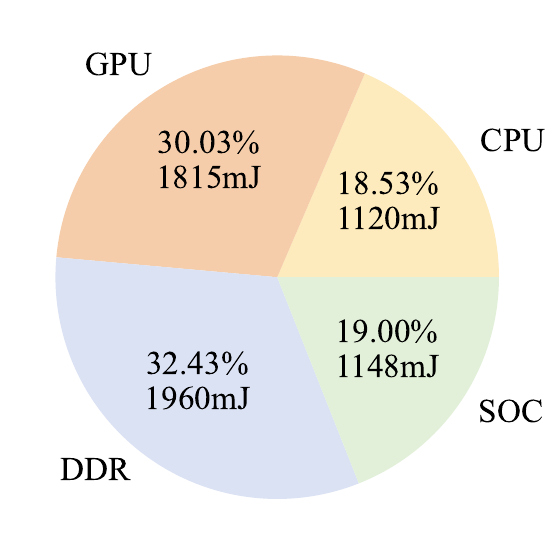}%
        }
        \hfil
        \subfloat[DeeoLabv3+.]{\includegraphics[width=0.24\textwidth]{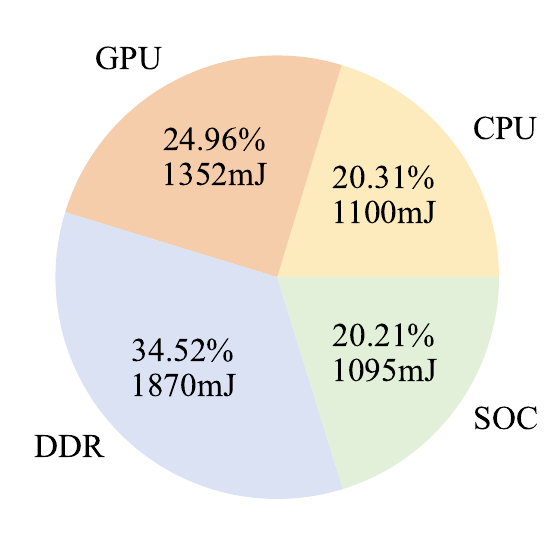}%
        }
        \hfil
        \subfloat[SkySeg (leader UAV).]{\includegraphics[width=0.24\textwidth]{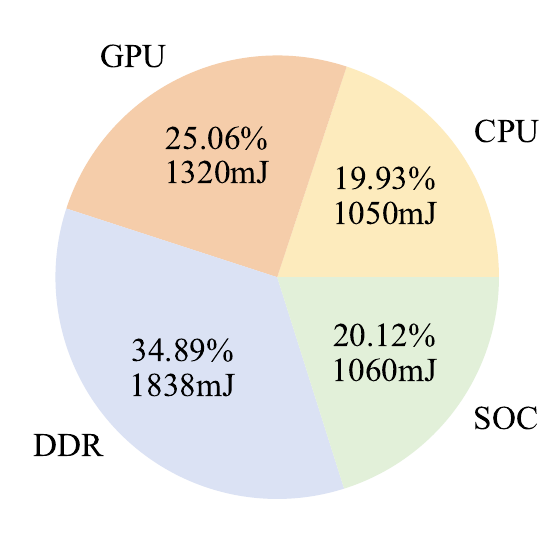}%
        }
        \hfil
        \subfloat[SkySeg (follower UAV).]{\includegraphics[width=0.24\textwidth]{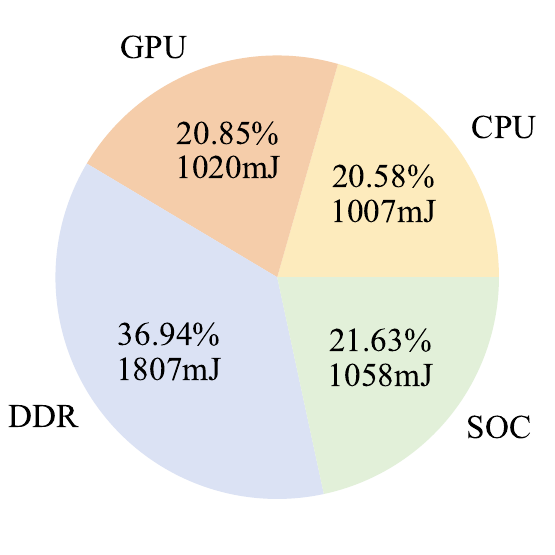}%
        }
        \caption{Average power consumption of each module on the TX2 for the SDD dataset.}
	\label{fig:energy}
\end{figure*}

\begin{figure*}[ht]
	\centering
        \subfloat[SegFormer-B1.]{\includegraphics[width=0.24\textwidth]{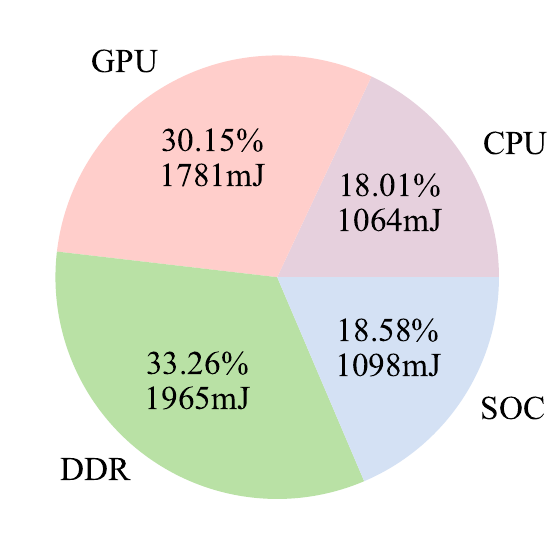}%
        }
        \hfil
        \subfloat[DeeoLabv3+.]{\includegraphics[width=0.24\textwidth]{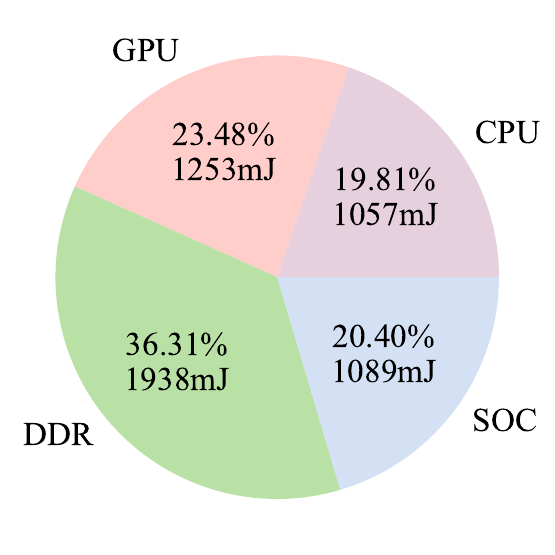}%
        }
        \hfil
        \subfloat[SkySeg (leader UAV).]{\includegraphics[width=0.24\textwidth]{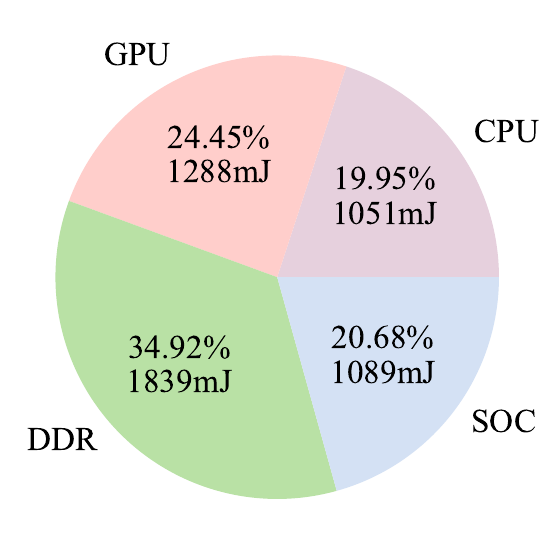}%
        }
        \hfil
        \subfloat[SkySeg (follower UAV).]{\includegraphics[width=0.24\textwidth]{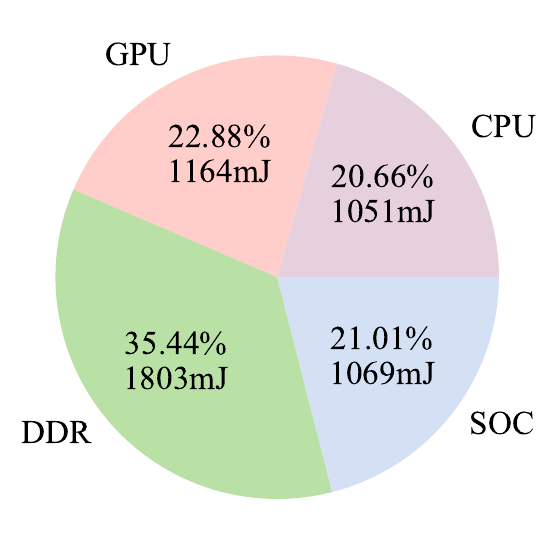}%
        }
        \caption{Average power consumption of each module on the TX2 for the FloodNet dataset.}
	\label{fig:energy1}
\end{figure*}

For SDD-C, each follower UAV transmits predicted class labels (600 $\times$ 400 integers) and prediction probabilities (floats) to the leader UAV. With integers occupying 2B and floats 4B, the total data volume is 600 $\times$ 400 $\times$ (2+4) = 1.4MB. For FloodNet-C, the transmitted data is 400 $\times$ 300 $\times$ (2+4) = 0.72MB.
For cross-device TTA, for the L-F setup, each UAV updates statistics locally. In L-2F, the leader UAV updates its statistics locally, while the two follower UAVs compute and share their statistics. In L-3F, each follower UAV transmits statistics to the other two follower UAVs.
The DeepLabv3+ model on follower UAVs includes 60 BN layers with 17872 channels in total. Each BN layer transmits two floating-point values (mean and variance) per channel, resulting in 17872 $\times$ (2+2) = 71.488KB of data per follower UAV. 

\subsection{TX2 Power Analysis}
We use the NVIDIA Jetson TX2 system resource monitor, jtop, to test power consumption during onboard computation on the UAV, which records power changes for each module twice per second. The mainstream semantic segmentation models we compare, as well as our SkySeg, are all deployed on the TX2. The average power consumption of each module on the TX2 during inference is shown in Fig.~\ref{fig:energy} and Fig.~\ref{fig:energy1}. GPU power consumption exhibits the most significant differences across methods. 
Additionally, models for single UAVs equipped with high-cost sensors typically require over 100 GFLOPs of computation, making them impractical for resource-constrained micro UAVs. In contrast, our SkySeg stays under 60 GFLOPs per device, significantly reducing computational overhead and enabling deployment across various resource-limited UAV platforms.

\section{Conclusion}
This paper presents SkySeg, a heterogeneous multi-UAV cooperation framework designed for efficient onboard semantic segmentation using low-cost sensors. It addresses hardware constraints and data acquisition challenges on UAVs by integrating collaborative perception with cross-device TTA. To enhance segmentation accuracy, SkySeg proposed an information fusion inference method that combines low-definition, wide-area images with high-definition, focused-area images. To adapt to dynamic environments, SkySeg employs cross-device TTA among follower UAVs. Experimental results demonstrate that SkySeg improves segmentation accuracy by 5.91\% on clean SDD dataset and 10.91\% on corrupted data while reducing inference latency by about 3.6x. SkySeg offers an efficient and adaptive solution for UAV semantic segmentation infield. By integrating computer vision and cooperative flight pattern, our SkySeg framework provides new insights for autonomous UAV perception and decision-making. It also serves as an important reference for the development of multi-UAV collaborative perception systems.

\nocite{*}
\bibliography{reference.bib} %bibfile_name
\bibliographystyle{IEEEtran}

\end{document}